\documentclass[conference]{IEEEtran}
\IEEEoverridecommandlockouts
\usepackage{cite}
\usepackage{amsmath,amssymb,amsfonts}
\usepackage{algorithmic}
\usepackage{graphicx}%, subcaption}
\usepackage{textcomp}
\usepackage{xcolor}
\usepackage{tikz}
\usepackage{hyperref}
\usepackage{pgfplots}
\pgfplotsset{
    compat=newest
}

\def\BibTeX{{\rm B\kern-.05em{\sc i\kern-.025em b}\kern-.08em
    T\kern-.1667em\lower.7ex\hbox{E}\kern-.125emX}}
\begin{document}

\title{A Deep Learning Model for Chilean Bills Classification\\
}

\author{\IEEEauthorblockN{Daniel San Mart\'{i}n}
\IEEEauthorblockA{\textit{Departamento de Inform\'{a}tica} \\
\textit{Universidad T\'{e}cnica Federico Santa Mar\'{i}a}\\
Valpara\'{i}so, Chile \\
daniel.sanmartinr@usm.cl}
\and
\IEEEauthorblockN{Daniel Manzano}
\IEEEauthorblockA{\textit{Unidad de An\'{a}lisis Institucional y Datos} \\
\textit{Universidad de Chile}\\
Santiago, Chile \\
danielmanzano@uchile.cl}
%\and
}

\maketitle

\begin{abstract}
    Automatic bill classification is an attractive task with many 
    potential applications such as automated detection and counting 
    in images or videos. To address this purpose we present
    a Deep Learning Model to classify Chilean Banknotes, 
    because of its successful results in 
    image processing applications. For optimal performance of
    the proposed model, data augmentation techniques are introduced
    due to the limited number of image samples.
    Positive results were achieved in this work, verifying 
    that it could be a stating point to be extended to more complex applications.
\end{abstract}


\section{Introduction} \label{sec:introduction}

    The automatic classification of bills may be an interesting work
    as previous step for more complex applications for institutions such as 
    banks or casinos. They need quick and accurate control and sorting operations 
    to handle large amount of cash. Since each bill has its own design, human 
    vision can distinguish them easily by their letters, colors and patterns, but 
    for a machine vision system this would be slightly difficult. For computers 
    algorithms, it is necessary to obtain the highest number of image characteristics 
    to identify each bill design. \emph{Deep Neural Networks (DNN)} have achieved exceptional results on image 
    feature extraction. They have demonstrated to be useful for objects and faces 
    detection even introducing different light conditions and object orientations \cite{Lecun2015}. 
    Also, in recent years, \emph{DNN} has allowed the detection of simultaneous objects. 
    In particular, \emph{Convolutional Neural Networks (CNN)} are the most used model 
    for image processing \cite{Rawat2017} and applications such counterfeit bill detection or bills 
    portrait detection \cite{Kitagawa2017, GaleanaPerez2018, Lee2018, Lee2018a}. %There is some work in this area applied to other countries 
    %bills such as those presented in \cite{Kitagawa2017, GaleanaPerez2018, Lee2018, Lee2018a}.
    %where the main idea is to identify the image patterns.
    In this work we propose a \emph{CNN} model to classify images of Chilean banknotes. 
    Due to the reduced number of samples, we describe the use of different data 
    augmentation techniques to artificially increase the number of images and 
    improve the classification process. Good results are achieved 
    for a sufficient amount of data, without the need to excessively increase the 
    complexity of the model.

\section{Methods} \label{sec:methods}

    \subsection{Data Collection} \label{subsec:data_collection}
    
        Chile has $5$ different banknotes for $1000$, $2000$, $5000$, $10000$ and $20000$ 
        pesos \cite{BancoCentral2019}. To collect the data, we download both front and 
        back bills images provided by the Central Bank of Chile website. Since a \emph{DNN} 
        approach for image classification needs a large amount of samples, we need to increase
        the dataset size using techniques for data augmentation.
    
    \subsection{Dataset Augmentation} \label{subsec:dataset_augmentation}
    
\begin{figure}[!ht]
        \centering
        \includegraphics[width=.6\columnwidth]{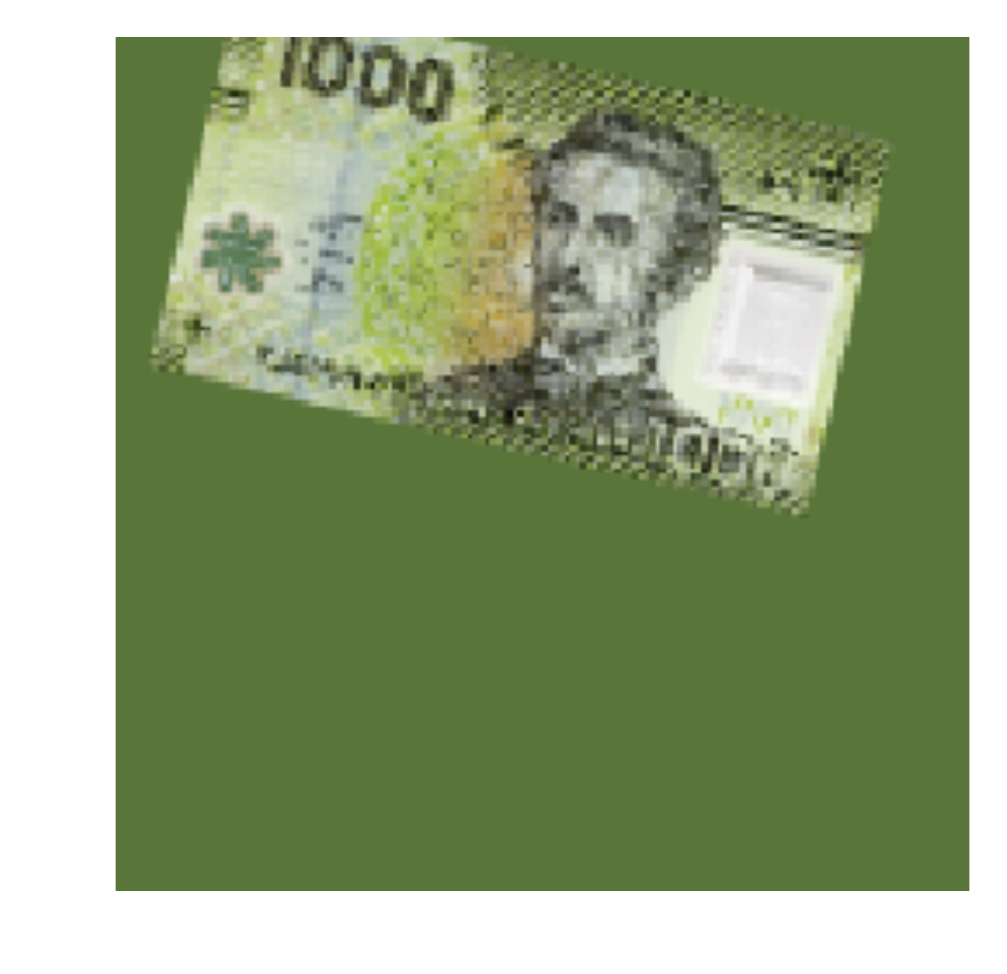}
        \includegraphics[width=.6\columnwidth]{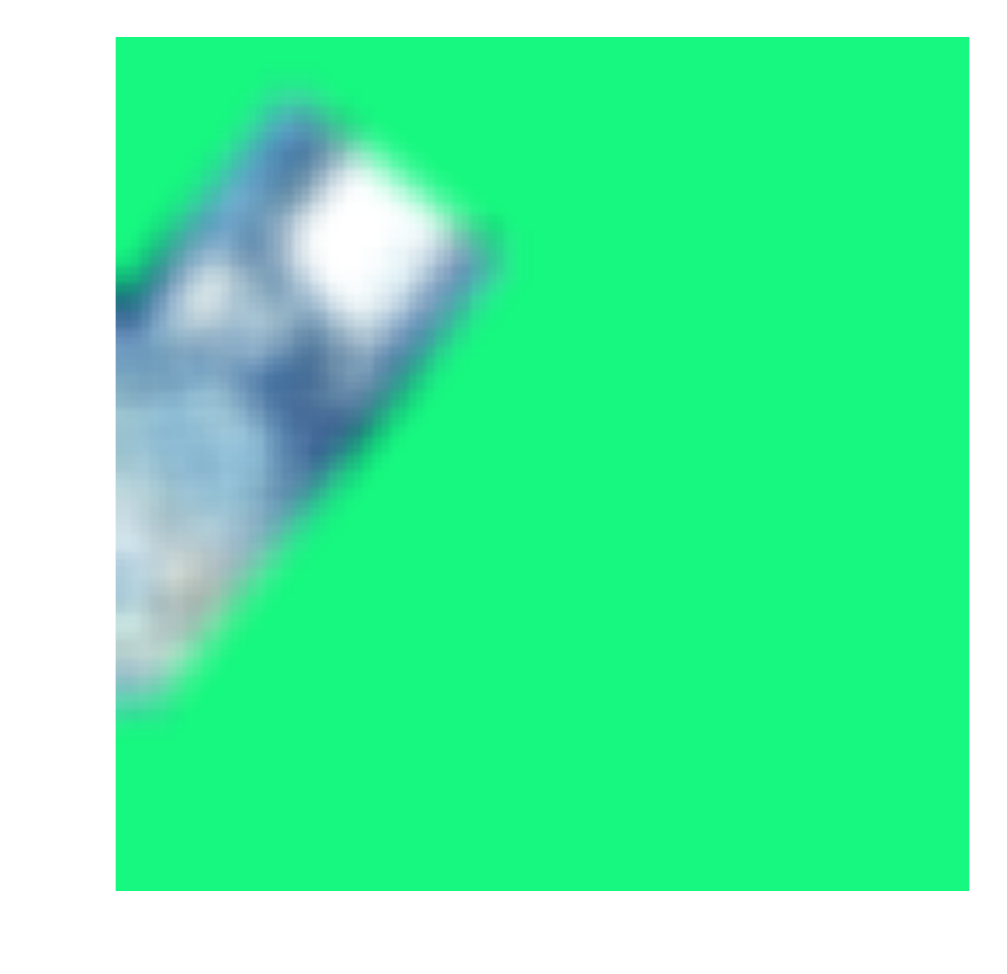}
        \includegraphics[width=.6\columnwidth]{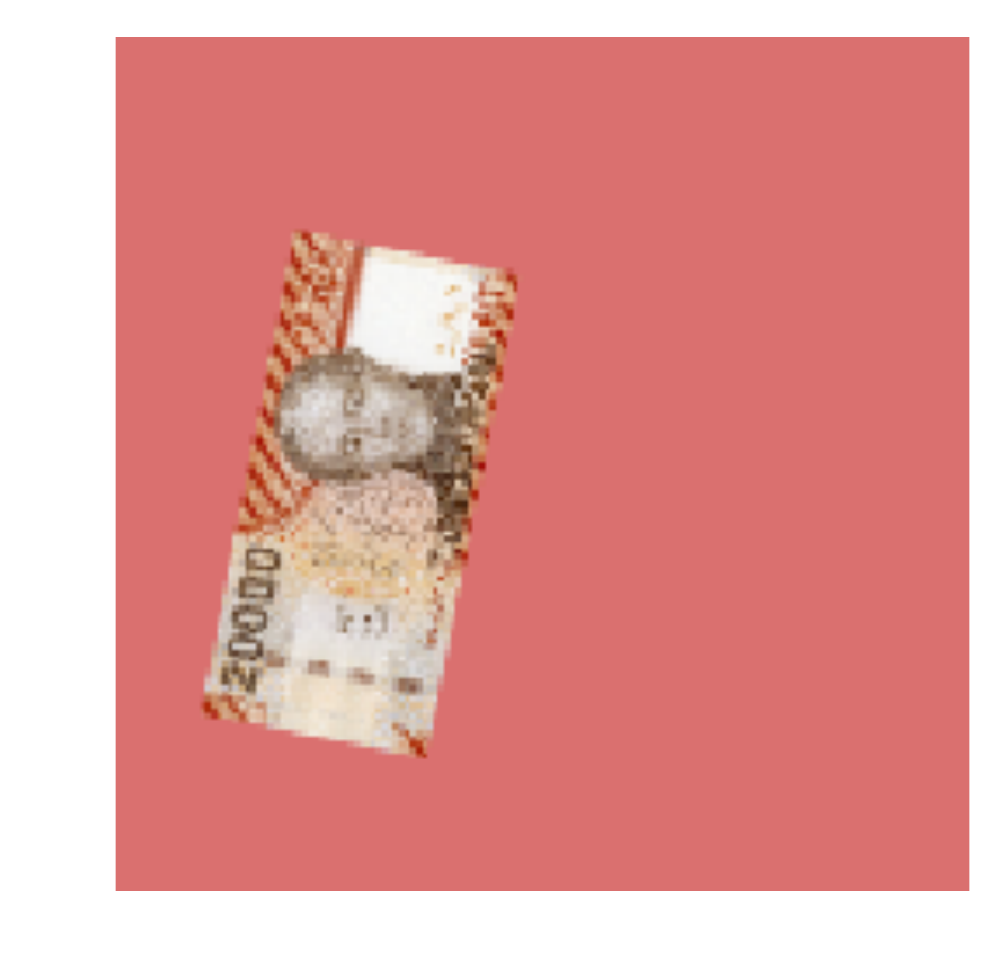}
        \caption{Three sample bills with data augmentation techniques. Above \$$1000$ banknote,
            middle \$$10000$ banknote and below \$$20000$ banknote.}
        \label{fig:samples}
    \end{figure}

        To enhance image classification task, \emph{DNN} are trained on massive number of data 
        examples, but in some areas, like medical imaging, data is scarce or expensive to generate, 
        therefore, augmentation techniques are used to solve this problem \cite{Goodfellow-et-al-2016}. 
        Moreover, \emph{DNN} have a large number of parameters to set, which usually causes \emph{overfitting}. Data 
        augmentation acts as a \emph{regularizer} to deal this issue. On the other hand it is important 
        to choose good data augmentation functions, because they can slow down training and can 
        introduce biases into the dataset \cite{Cubuk2018}. Some of the techniques used in this work were: translations, 
        rotations, scaling, brightness variation, Gaussian blur, random figures over the image, etc.
        Fig.~\ref{fig:samples} presents samples of generated images.

    \subsection{Convolutional Neural Networks} \label{subsec:convnet}
    
        \emph{CNN} are simply neural networks that use convolution in place of general 
        matrix multiplication in at least one of their layers \cite{Goodfellow-et-al-2016}. This
        neural networks consists of an input layer, one or more convolution layers, one or many fully 
        connected layer, and an output layer. \emph{The convolution layer} generally includes a convolution 
        operation, a pooling operation, and an activation function. The \emph{Convolution} is an operation of 
        a matrix of weights called \emph{Kernel} over the image data matrix. \emph{The Kernel} 
        slides over the image inputs, performing an element-wise multiplication and then summing 
        up the results into a single output pixel. It works like a filter and the repeated application 
        of the same filter to an input results in a map of activations called a \emph{Feature map}. 
        \emph{The Pooling operation} reduces the size of the image by leaving only values that 
        satisfy certain rules among the pixels in a specific area \cite{Lee2018}. Usually \emph{max} or 
        \emph{average} is used. \emph{The Activation Function} is used to apply nonlinearity to the results of the previous layer. 
        Generally \emph{ReLu}, \emph{Sigmoid} and \emph{Hyperbolic Tangent} functions are used \cite{Lee2018}.
        \emph{The fully connected layer} is a simple feed forward neural networks and is the most basic 
        component of \emph{CNN}. This layer form the last few layer in this network. The input of this layer is 
        the flattened output from the final pooling or convolutional layer, feeding 
        into the fully connected layer. For classification tasks, \emph{Softmax} is used as activation function
        in the output layer.
        Other methods included in this kind of models are \emph{Dropout}, technique used to reduce \emph{overfitting} 
        removing neural units randomly \cite{Srivastava2014}; and \emph{Batch Normalization} used to improve 
        optimization process normalizing layers inputs \cite{Ioffe2015}.
     
    \subsection{AlexNet} \label{subsec:alexitico}
    
        \emph{AlexNet} is a Deep Model architecture used in ImageNet Large Scale Visual Recognition Challenge \cite{Krizhevsky2017}.
        %This model is computationally expensive, but take advantage of the use of GPU computation for training process.
        In our proposed model we keep the original design of \emph{AlexNet} only setting the number of neurons and
        size of convolution kernels. \emph{AlexNet} has $8$ layers, the first $5$ are convolutional layers,
        and the last $3$ are fully connected. The detailed configuration used was:
        \begin{enumerate}
            \item Convolution layer with $32$ filters, kernel size of $(11, 11)$ and $(2,2)$ \emph{max pooling}.
            \item Convolution layer with $64$ filters, kernel size of $(5, 5)$ and $(2,2)$ \emph{max pooling}.
            \item Convolution layer with $128$ filters, kernel size of $(3, 3)$ and $(2,2)$ \emph{max pooling}.
                Also include a zero-padding of size $(1,1)$.
            \item Convolution layer with $256$ filters, kernel size of $(3, 3)$ and zero-padding of size $(1,1)$.
            \item Convolution layer with $256$ filters, kernel size of $(3, 3)$ and $(2,2)$ \emph{max pooling}.
                Also include a zero-padding of size $(1,1)$.
            \item Dense layer with $512$ neurons and \emph{dropout} of $50$\%.
            \item Dense layer with $32$ neurons and \emph{dropout} of $50$\%.
            \item Dense layer with $5$ neurons.
        \end{enumerate}
        
        The activation function used in all layer was \emph{ReLu} except last one with \emph{Softmax}. Also, all
        layers used \emph{Batch Normalization}.
        Training process were performed using \emph{Stochastic Gradient Descent} with $0.1$ of \emph{learning rate},
        using $100$ samples as \emph{batch size} and $400$ epochs.

    \subsection{Implementation} \label{subsec:implementation}
            
        The model was implemented with \emph{Python} using the following libraries:
        \begin{itemize}
            \item \emph{NumPy} for data structures manipulation.
            \item \emph{OpenCV} for image processing, including data augmentation techniques. 
            \item \emph{Scikit-Learn} for metrics and data processing.
            \item \emph{Keras}, a library to build \emph{Machine Learning} models avoiding too much code.
        \end{itemize}
        
        Experiments were performed in computational infrastructure provided by \emph{Leftraru cluster}
        from the \emph{National Laboratory of High Performance Computing Chile (NLHPC)}.
        % The first stage was the process of dataset augmentation and preparation. Due to the small number of
        % samples we apply a set of random data augmentation techniques and artificially increase the number 
        % of images in the dataset. Then, we split the dataset into $2/3$ of training and $1/3$ of testing set.
        % The training set is also split into $2/3$ of real training set and $1/3$ of validation set for
        % metrics evaluation in the model training step.

\section{Results} \label{sec:results}

    In this section we first define the metrics used in the model evaluation and then we show 
    the results obtained.
    %This section we presents the results of the model proposed. First we define the metrics used to evaluate
    %the model performance.
    \subsection{Metrics} \label{subsec:metrics}
    
        Let $M$ a matrix where $i$-th row represents the true class label and $j$-th column the predicted class label.
        Each element $M_{i,j}$ is the number of times that the classifier predicted label $i$ as $j$. 
        
        \begin{itemize}
            \item \emph{Precision} is the number of correct predictions of class $i$ 
                versus the total of values predicted as $i$.
                %The precision is intuitively the ability of the classifier not to label as positive a sample that is negative
                \begin{equation*}
                    Precision_i = \frac{M_{i,i}}{\sum_{j} M_{j,i}}
                \end{equation*}
            
            \item \emph{Recall} is the number of correct predictions of class $i$, versus total 
                number of instances that should have label $i$.
                \begin{equation*}
                    Recall_i = \frac{M_{i,i}}{\sum_{j} M_{i,j}}
                \end{equation*}
                
            \item $F_1$ \emph{score} integrate precision and recall respective to a specific positive class. 
                The $F_1$ score can be interpreted as a weighted average of the precision and recall.%, where an 
                %$F_1$ score gets its best value at $1$ and worst at $0$.
                \begin{equation*}
                    F_1 = 2\,\frac{Precision \times Recall}{Precision + Recall}
                \end{equation*}
                
            \item \emph{Accuracy} is the ratio between number of correct predictions over the total 
                of predictions made.% number of input samples.
                \begin{equation*}
                    Accuracy = \frac{\sum_{i} M_{i,i}}{\sum_{i}\sum_j M_{i,j}}
                \end{equation*}
        \end{itemize}

    \subsection{Model performance} \label{subsec:model_performance}

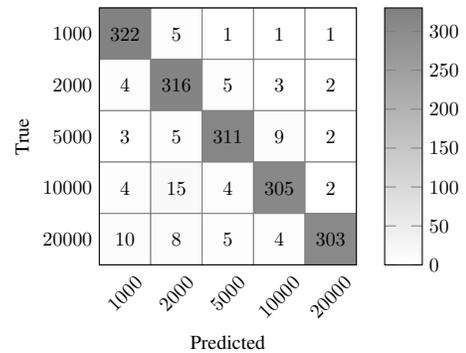
\begin{figure}[!ht]
        \centering
        \begin{tikzpicture}[scale=1, every node/.style={scale=0.75}]
    \begin{axis}[
        width=5cm,
        height=5cm,
        colormap={whiteblack}{color(0cm)  = (white);color(1cm) = (gray)},%{bluewhite}{color=(white) color=(blue)},%rgb255=(0,0,139)},
        xlabel={Predicted},
        ylabel={True},
        xticklabels={$1000$, $2000$, $5000$, $10000$, $20000$},
        xtick={0,1,2,3,4},
        xtick style={draw=none},
        yticklabels={$1000$, $2000$, $5000$, $10000$, $20000$},
        ytick={0,1,2,3,4},
        x tick label style={rotate=45},
        ytick style={draw=none},
        enlargelimits=false,
        colorbar,
        colorbar style={
        plot graphics/node/.style={scale=1.33,anchor=south west,inner sep=0pt,},
            ytick={0,50,100,150,200,250,300},
            yticklabel={\pgfmathprintnumber\tick},
            yticklabel style={
                    /pgf/number format/fixed,
            /pgf/number format/precision=1}
        },
        point meta min=0.0,point meta max=330.0,
        nodes near coords={\pgfmathprintnumber\pgfplotspointmeta},
        nodes near coords style={
                yshift=-7pt,
        /pgf/number format/fixed,
                /pgf/number format/precision=2},
    ]
        \addplot[
            matrix plot,
            mesh/rows=5,
            mesh/cols=5,
            point meta=explicit,draw=gray
        ] table [meta=C] {
            x y C
            0 0 322
            1 0 5
            2 0 1
            3 0 1
            4 0 1
            0 1 4
            1 1 316
            2 1 5
            3 1 3
            4 1 2
            0 2 3
            1 2 5
            2 2 311
            3 2 9
            4 2 2
            0 3 4
            1 3 15
            2 3 4
            3 3 305
            4 3 2
            0 4 10
            1 4 8
            2 4 5
            3 4 4
            4 4 303
        };
    \end{axis}
\end{tikzpicture}
        \caption{Confusion Matrix with classification results.}
        \label{fig:con_mat}
    \end{figure}

        To measure model performance we generate a dataset of $5000$ samples, with $1000$ samples per class.
        The images where scaled to a fixed size of $150\times 150$ using the $3$ RGB channels. $50\%$ of the images was 
        front and $50\%$ back for each class and also a $10\%$ of threshold was used to apply a random \emph{data augmentation} technique
        described in sub-section~\ref{subsec:dataset_augmentation}.
        Then, dataset was split into \emph{training} and \emph{testing} sets in a proportion of $67\%$ and $33\%$ respectively.
        \emph{Training} set was also split into the same proportion for a \emph{validation} set to analyze metrics in training step.
        Fig.~\ref{fig:con_mat} shows the \emph{Confusion Matrix} associated with the experiment carried out.
        Visually we notice that the classifier has a satisfactory performance, making mistakes only in a few samples. 
        Using metrics described before and with \emph{Confusion Matrix} as $M$, Table~\ref{tab:results} summarizes
        the model performance. Numerically we validate that our proposal reaches more than $90\%$ in the metrics presented
        and an accuracy of $94\%$.

\section{Contribution} \label{sec:contribution}
    
    Interesting results were achieved by means of a \emph{CNN} model based
    on \emph{Alexnet} architecture and data augmentation techniques.
    This study provides a proof of concepts as a first step to make more complex applications 
    such as bill recognition and counting. Focusing on casinos or banks problems, this classification
    task gives the possibility of quick bills process through images or videos. 
    %This approach could reduce humans mistakes when bills amount are too large and in order to enhance this task we can use machines to be more accurate performing this kind of work.
    When the number of banknotes is very large, we could use this automatic approach to improve 
    performance in this type of task and therefore help to reduce the number of human mistakes.

    All the source code used in this work are online in a public repository and may be accessed 
    in \url{https://github.com/dsanmartin/chilean-bills-classification}.
    %\This work is a little proof of concept for more complex applications
    %\such as bill recognition and counting.

\section*{Acknowledgment}

    This research was partially funded by CONICYT-PFCHA/Doctorado Nacional/2019-21191017 
    and with the support of the \emph{Departamento de Inform\'{a}tica}, UTFSM, Chile.
    Powered@NLHPC: This research/thesis was partially supported by the supercomputing infrastructure of the NLHPC (ECM-02)

\bibliographystyle{IEEEtran}
\bibliography{IEEEabrv,references}

\appendix

    % Results
    \begin{table}[!ht]
        \caption{Classification Results}
        \centering
        \begin{tabular}{|c|r|r|r|}
            \hline
            \textbf{}&\multicolumn{3}{|c|}{\textbf{Metrics}} \\
            \cline{2-4} 
            \textbf{Bill Class} & \textbf{\textit{Precision}}& \textbf{\textit{Recall}}& \textbf{\textit{f1-score}} \\
            \hline
            \$ $1000$ & $0.939$ & $0.976$ & $0.957$ \\
            \$ $2000$ & $0.905$ & $0.958$ & $0.931$ \\
            \$ $5000$ & $0.954$ & $0.942$ & $0.948$ \\
            \$ $10000$ & $0.947$ & $0.924$ & $0.936$ \\
            \$ $20000$ & $0.977$ & $0.918$ & $0.947$ \\
            & & & \\
            Micro Average & $0.944$ & $0.944$ & $0.944$ \\
            Macro Average & $0.945$ & $0.944$ &  $0.944$ \\
            Weighted Average & $0.945$ & $0.944$ & $0.944$ \\
            \hline
            %\multicolumn{4}{l}{$^{\mathrm{a}}$Sample of a Table footnote.}
        \end{tabular}
        \label{tab:results}
    \end{table}
    
    % Samples

    % Confusion matrix

\end{document}